\documentclass[final]{cvpr}

\usepackage{times}
\usepackage{epsfig}
\usepackage{graphicx}
\usepackage{amsmath}
\usepackage{amssymb}
\usepackage{xcolor}
\usepackage{booktabs}
\usepackage{soul}
\usepackage{multirow}

\usepackage[pagebackref=true,breaklinks=true,colorlinks,bookmarks=false]{hyperref}

\def\cvprPaperID{11} % *** Enter the CVPR Paper ID here
\def\confYear{CVPR 2021}
\begin{document}

%%%%%%%%% TITLE
\title{FReTAL: Generalizing Deepfake Detection using Knowledge Distillation and Representation Learning}

\author{Minha Kim\thanks{equal contribution} , Shahroz Tariq\footnotemark[1]\\
College of Computing and Informatics\\
Sungkyunkwan University, South Korea\\
{\tt\small \{kimminha,shahroz\}@g.skku.edu}
% \and
% Shahroz Tariq\\
% Sungkyunkwan University, South Korea\\
% {\tt\small shahroz@g.skku.edu}

\and
Simon S. Woo\thanks{corresponding author}\\
Department of Applied Data Science\\
Sungkyunkwan University, South Korea\\
{\tt\small swoo@g.skku.edu}
}

\maketitle
%\thispagestyle{empty}

%%%%%%%%% ABSTRACT
\begin{abstract}
As GAN-based video and image manipulation technologies become more sophisticated and easily accessible, there is an urgent need for effective deepfake detection technologies. Moreover, various deepfake generation techniques have emerged over the past few years. While many deepfake detection methods have been proposed, their performance suffers from new types of deepfake methods on which they are not sufficiently trained.  To detect new types of deepfakes, the model should learn from additional data without losing its prior knowledge about deepfakes (catastrophic forgetting), especially when new deepfakes are significantly different. In this work, we employ the Representation Learning (ReL) and Knowledge Distillation (KD) paradigms to introduce a transfer learning-based Feature Representation Transfer Adaptation Learning (FReTAL) method. We use FReTAL to perform domain adaptation tasks on new deepfake datasets, while minimizing the catastrophic forgetting. Our student model can quickly adapt to new types of deepfake by distilling knowledge from a pre-trained teacher model and applying transfer learning without using source domain data during domain adaptation. Through experiments on FaceForensics++ datasets, we demonstrate that FReTAL outperforms all baselines on the domain adaptation task with up to 86.97\% accuracy on low-quality deepfakes.

\end{abstract}
\section{Introduction}
\label{sec:Intro}

% \sh{Deep learning is booming}

% \sh{but some people use it for bad purpose like deepfake}

% \sh{include citations}

Synthetic multimedia is becoming increasingly common on the Internet and social media~\cite{DeepfakeSocialMedia1,DeepfakeSocialMedia2}. Its popularity is being driven by the widespread availability of simple tools and techniques for creating realistic fake multimedia information~\cite{Deepfakes,FaceApp,FaceSwap}. Recent advances in deep learning have aided in the generation methods for creating synthetic images and videos that look remarkably close to real-world images~\cite{PGGAN,StyleGAN,StarGAN} and videos~\cite{VideoGAN_DVDGAN,VideoGAN_TGAN-f,VideoGAN_TGANv2}. Especially, deepfakes are manipulated multimedia generated using such techniques, which generally involve neural networks such as Autoencoders (AE)~\cite{DeepLearningBook} and Generative Adversarial Networks (GAN)~\cite{GANPaper}. Although such tools can help in automating game design~\cite{DeepfakeGameDesign}, photorealistic scenery generation~\cite{GauGAN}, film making~\cite{DeepfakeFilmMaking}, human face generation~\cite{PGGAN} or virtual and augmented reality rendering~\cite{DeepfakeARVR}, they can also be very dangerous and misused if utilized for malicious purposes~\cite{news1,news2,news3,news4}. The line between real and fake media is becoming increasingly blurred as manipulation techniques become more available, practical, and difficult to be detectable.

\begin{figure}
    \centering
    \includegraphics[clip, trim=15pt 0pt 15pt 0pt,width=1\linewidth]{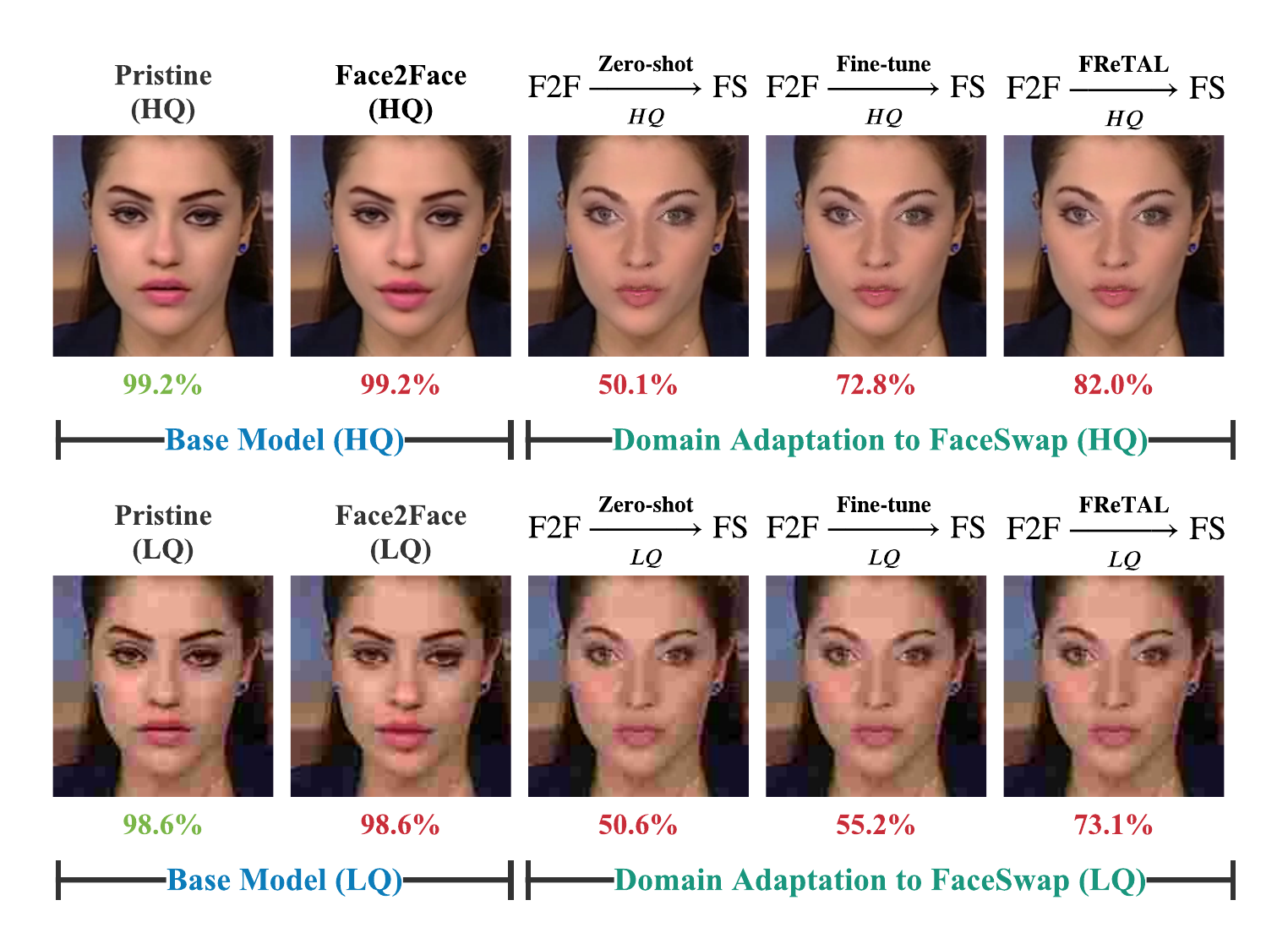}
    \caption{\textbf{Performance of domain adaptation task.} We use Face2Face (F2F) as the source and FaceSwap (FS) as target dataset. Xception is used as the backbone model. Here, we demonstrate two cases: 1) high-quality deepfakes (\textit{top}) for source and target. 2) low-quality (\textit{bottom}) deepfakes for source and target. Our FReTAL performs significantly better than zero-shot and fine-tuning on low quality deepfakes (more comparison in Section~\ref{sec:Results}).}
    \label{fig:motivation}
\end{figure}

For example, using face swaps-based deepfakes, an attacker can put the victims in place and setting where they have never been. This type of deepfakes is easily used to generate pornographic videos of celebrities and masses alike~\cite{news_deepfake96percent}. Also, by manipulating the lip movement with facial reenactment techniques and the associated speech signal, deepfake videos of people speaking words that they never said can be produced. For example, many fake propaganda videos are generated by reanimating real videos of political figures~\cite{ObamaDeepfake,TrumpClimateDeepfake,NixonDeepfake}.

Among the most famous deepfake forgeries are human facial manipulations~\cite{FaceForensics++,Face2Face,FaceSwap,Deepfakes,NeuralTextures,PGGAN}. However, due to a lack of data, detecting these deepfakes or forged images/videos is challenging in a real-world setting. To increase data availability, the research community has recently released a slew of deepfake datasets to assist other researchers around the world in developing detection mechanisms for such deepfakes. The FaceForensics++~\cite{FaceForensics++} dataset is one of the earliest and most popular benchmark deepfake datasets. Many other deepfake benchmark datasets are recently released, such as Deepfake Detection Challenge~\cite{DFDC} from Facebook and CelebDF~\cite{CelebDF}.

% \sh{Nonetheless, these benchmark datasets facilitate enhancing performance and diversifying detection approaches. Many deepfake detection methods achieve high test accuracy on a single deepfake dataset. However, they perform poorly on deepfakes created from novel methods that were not introduced during the training process. In other words, they lack generalizability. As a result, it is normal that such approaches perform poorly in the real world, especially on low-quality deepfake videos. Also, as investigated by Tariq et al., developing a generalized classifier that performs well uniformly on various types of deepfakes is of paramount importance. And not much research is done in that direction. Furthermore, it would also be unrealistic to generate a large dataset for each novel deepfake generation method to train a deepfake detector. Therefore, a more widely applicable solution to the problem of deepfake detection is required. In this work, we aim to explore such a solution by keeping data scarcity and domain adaptation into consideration.}

As a result, these benchmark datasets facilitate improving performance and diversifying detection approaches. Many deepfake detection methods achieve high test accuracy on a single deepfake dataset~\cite{FaceForensics++,MesoNet,Shahroz1}. However, they perform poorly on deepfakes created from novel methods that were not introduced during the training process~\cite{Shahroz3}. In other words, they lack generalizability, especially on low-quality (compressed) deepfake videos, which are the focus of our paper as deepfakes shared on social media typically goes through compression. Therefore, such approaches trained on high quality videos generally perform poorly in the real world.

The successful demonstration of the deepfake impersonation attack on commercial and open-source face recognition and authentication APIs by Tariq et al.~\cite{ShahrozAPI} highlights the importance of developing a generalized classifier that consistently performs well on various types of deepfakes. In particular, it is of paramount importance in the face authentication domain, and not much research has been conducted in that direction. Furthermore, it would also be unrealistic to generate a large dataset tailored toward each deepfake generation method to train a deepfake detector. Therefore, a more widely applicable generalized solution to the problem of deepfake detection is required. In this work, we aim to explore such an approach by keeping data scarcity and domain adaptation into consideration. \textit{Note: it is relatively easy for the model to detect high-quality (uncompressed) deepfakes, as shown by previous research}~\cite{Shahroz3,FaceForensics++,MesoNet}, \textit{thereby we focus more on low-quality deepfakes than high-quality in our work.}

%\sh{Therefore, in this paper, keeping data scarcity and domain adaptation into consideration, we explore the solution in Knowledge Distillation (KD) and Representation Learning (RL) domains.}

In recent years, several Knowledge Distillation (KD)-based methods are proposed for domain adaptation tasks~\cite{KD_domainAdaptation4,KD_domainAdaptation5,KD_domainAdaptation3,KD_domainAdaptation2,KD_domainAdaptation1}. However, none of them have studied KD for domain adaptation in the media forensics domain, especially for deepfake detection. To this end, we propose Feature Representation Transfer Adaptation Learning (FReTAL), a Knowledge Distillation-based method for deepfake detection using representation learning (ReL) to transfer representations between the source (teacher) and target (student) domains.

% \sh{deepfake is causing so much privacy issues.}

% \sh{recent deepfake detectors lacks generalizability especially at low quality}

% \sh{In this paper we propose FReTAL}

The main contributions of our work are summarized as follows:

\begin{enumerate}
    \item We propose a novel domain adaption framework, Feature Representation Transfer Adaptation Learning (FReTAL), based on knowledge distillation and representation learning that can prevent catastrophic forgetting without accessing the source domain data.
    \item We show that leveraging knowledge distillation and representation learning can enhance adaptability across different deepfake domains.
    \item We empirically demonstrate that our method outperforms baseline approaches on deepfake benchmark datasets with up to 86.97\% accuracy on low-quality deepfakes.
\end{enumerate}

Our code is available here\footnote{\url{https://github.com/alsgkals2/FReTAL}}. The rest of this paper is organized as follows. We discuss related work of deepfake detection, KD, and ReL in Section~\ref{sec:Related}. We explain our FReTAL in Section~\ref{sec:Approach} and describe our experimental settings in Section~\ref{sec:Experiment}. Section~\ref{sec:Results} presents the results, and Section~\ref{sec:Discussion} provides a discussion and limitations of our work. Finally, we offer our conclusions in Section~\ref{sec:Conclusion}.
\section{Background and Related Work}
\label{sec:Related}
This work spans different fields, such as deepfake detection, domain adaptation, knowledge distillation, and representation learning. In this section, we will briefly cover the background and related research in aforementioned areas.

\subsection{Deepfakes}
While several sophisticated algorithms for creating realistic synthetic face videos have been developed in the past~\cite{Traditional_deepfake1,Traditional_deepfake2,Traditional_deepfake3,Face2Face,deepfake1,deepfake2,PGGAN,deepfake3,deepfake4,deepfake5,NeuralTextures}, most of these studies have not been widely available as open-source software applications so that the public can use. On the other hand, a much more straightforward approach focused on neural image style transfer~\cite{StyleTransfer1,StyleTransfer2} has emerged as the preferred method for creating deepfake videos on a large scale. Now, many open-source implementations are now publicly available in the form of FakeApp~\cite{FakeApp}, DeepFaceLab~\cite{DeepFaceLab}, FaceApp~\cite{FaceApp}, and many others~\cite{Deepfakes,FaceSwap}. Even though the core idea is the same, each method has a slightly different implementation, resulting in different types of deepfakes. And, these methods are continuously improving over the years.

\textbf{Deepfake Detection.}
As deepfakes have become a worldwide phenomenon, there has been a surge of interest in deepfake detection methods. The majority of current deepfake detection methods~\cite{Shahroz3,FaceForensics++,rossler2018faceforensics,MesoNet,ShahrozAPI,FaceXRay} rely on deep neural networks (DNNs). These methods include splice detection~\cite{DeepfakeDetection1,DeepfakeDetection2,DeepfakeDetection3,DeepfakeDetection4,DeepfakeDetection5,DeepfakeDetection6}, abnormal eye blinking~\cite{DeepfakeDetection7}, signal level artifacts~\cite{DeepfakeDetection8,DeepfakeDetection9}, irregular head poses~\cite{DeepfakeDetection10}, peculiar behavior patterns~\cite{DeepfakeDetection11,DeepfakeDetection12}, and many other data-driven methods that do not rely on particular traces or artifacts in the deepfake videos~\cite{Shahroz1,Shahroz2,CLRNet,SAMGAN,SAMTAR,HASAMCVPRW,transferlearning_tgd,Hyeonseong2,Hyeonseong3}. However, to the best of our knowledge, except for the work of Cozzolino et al.~\cite{cozzolino2018forensictransfer} and Tariq et al.~\cite{Shahroz3}, not much research is conducted to apply domain adaptation on deepfake detection tasks.  Furthermore, Tariq et al.~\cite{Shahroz3} used high-quality deepfakes, and Cozzolino et al.~\cite{cozzolino2018forensictransfer} used medium-level compression (c23) on deepfake videos for domain adaptation. Our work is different from these works in that we use high-level compression (low-quality) because such deepfake videos are most common on social media.

\subsection{Representation Learning}
Representation learning (ReL) is the process of learning representations of input data, usually by transforming or extracting features from it, making a task like classification or prediction easier to perform.
For feedforward networks, ReL is simply representing the hidden layers by applying some conditions to the learned intermediate features~\cite{DeepLearningBook}. 

\textbf{Transfer Learning and Domain Adaptation. }
Transfer learning and domain adaptation apply to the situations in which the information learned in one context (for example, distribution $P_1$) is used to enhance generalization in another setting (say, distribution $P_2$).
% Grachten and Chacon~\cite{transferlearning_l2sp_l2} used L2-SP regularizer independently for transferring knowledge in vision applications. Using the starting point as an initiation of a fine-tuning mechanism and as a guide in the regularizer reliably improves the performance.
% Taking advantage of this point, Jeon et al.~\cite{transferlearning_tgd} presented the transferable GAN detection framework (T-GD). They proposed combining L2-SP and self-training~\cite{selftraining} to outperform the high performance.
In domain adaptation (DA), a subcategory of transfer learning, we apply an algorithm trained on the source domain to a different but related target domain.  The source and target domains have the same feature space but different distributions in DA. In comparison, transfer learning encompasses cases where the target domain's feature space is different from the source feature space~\cite{DeepLearningBook}.
%Transfer learning is a machine learning approach that reuses a learned model to improve performance of target learners while maintaining the knowledge of source domain. %It is a method of learning by making small changes called ``Fine-tuning''.
% In situations where large amounts of dataset learning are required, the transfer learning technique can train by reusing the pre-trained model without re-learning from scratch~\cite{X}.

As deepfake video generation techniques are continuously evolving, more types of deepfake videos will emerge in the future. Collecting and producing a large number of new deepfake samples for each dataset would be impractical. In this work, we use feature-based domain adaptation to detect deepfakes generated using various methods. It also reduces the time cost. In order to perform domain adaptation, the model is initialized with the pre-trained weights on the source dataset. That model is then used to learn a new target dataset.
Furthermore, if the source and the target domains are similar, we can improve the performance over existing models. However, if they are not then, it can lead to catastrophic forgetting~\cite{catastrophic1,catastrophic2}. Catastrophic forgetting is the tendency of a DNN to entirely and abruptly forget previously learned information upon learning new information. We solve this problem by using knowledge distillation.

%Also, transfer learning can reduce the time cost of learning from scratch.
% Furthermore, if the data from the source data and the target data to be newly learned are similar, we can improve the performance over existing models.
% However, if previously learned data do not exist, the more the model learns the target model, the more domain shifting occurs and catastrophic forgetting~\cite{catastrophic1,catastrophic2} of information pre-trained from source domain occurs. 
% Catastrophic forgetting which means to lose previous information while transfer learning.

% \textbf{Domain Adaptation.}
% 우리는 같은 deepfake detection task이지만, 데이터셋이 다르기 때문에 도메인 적응력을 높이기위해
% Feature Adaptation. For adapting the data from multiple
% sources, learning a common feature subspace or
% representation where the projected source and target
% domain are with similar distribution is generally resulted.
% However, learning common subspace generally
% belongs to shallow methodology, which relies on
% pre-trained deep models for discriminative feature
% representation. It is challenging to unify the shallow
% model with the deep model. Also, the heterogeneity
% of data distribution makes it challenging to gain such
% generic feature.

\textbf{Domain Adaptation using Knowledge Distillation.} 
Hinton et al.~\cite{hinton2015distilling} propose Knowledge distillation (KD). It is a method to compress knowledge of a large model to a small model. The main idea is that the student model can mimic the knowledge of the teacher model. Inspired by mimicking the teacher model, Li et al.~\cite{li2017learning} propose ``Learning without forgetting''. It is a method to maintain the source domain's knowledge by applying the knowledge distillation loss while transferring knowledge to the target domain. By adopting the principle of \textit{rehearsal}~\cite{chen2019catastrophic}, Rebuffi et al.~\cite{rebuffi2017icarl} propose to stores the information of the source domain (i.e., storing exemplars) to overcome catastrophic forgetting in class-incremental learning using KD loss. However, it requires a large amount of memory storage to store the features of the source domain. This may lead to privacy breaches, such as inversion attacks. To prevent this, we propose a Representation Learning-based method that does not need to store or use source data in the model while transfer learning.

\section{FReTAL}
\label{sec:Approach}
In this section, we provide details about our Feature Representation Transfer Adaptation Learning (FReTAL) method, including our motivation, and the pre-processing details.

\begin{figure*}[t!]
    \centering
    \includegraphics[width=\linewidth]{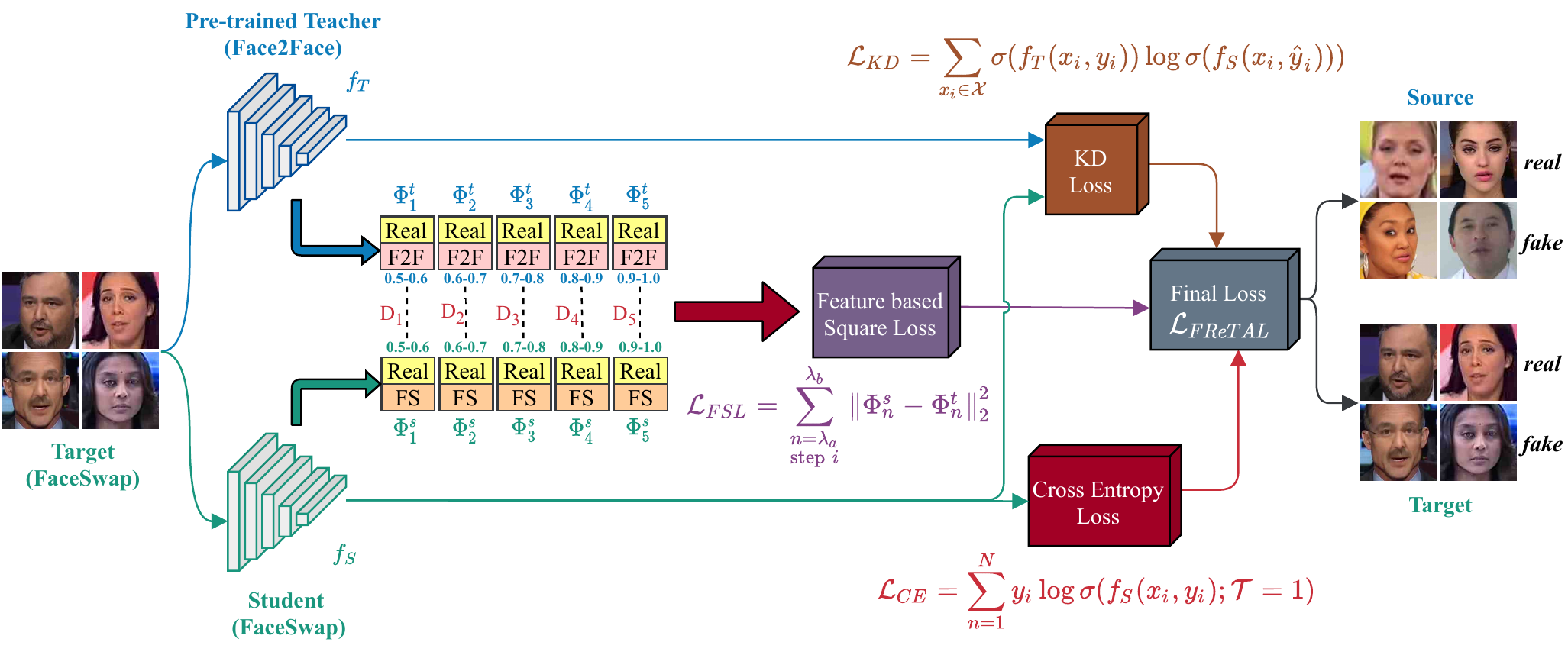}
    \caption{\textbf{The architecture of our Feature Representation Transfer Adaptation Learning (FReTAL)}. The teacher model is trained with the Xception.  Before transfer learning, we set the teacher model as untrainable. Then, we initialize the student model with the weights of the teacher model. Target domain data is provided to both teacher and student models to calculate the features for feature storage. We set the teacher as untrainable so that these features are fixed throughout the whole process. Whereas for the student, they will change in each iteration as training progress. We calculate KD loss between teacher and student models and a separate cross-entropy loss function just for the student model. Here, D1-D5 represents the square distance between feature storage of teacher and student.
    \textit{Note: for the first iteration of transfer learning, the teacher and student model will be the same.}}
    \label{fig:pipeline}
\end{figure*}

\textbf{Motivation. } Catastrophic forgetting is a big hurdle during domain adaptation tasks~\cite{Catastrophic3,Catastrophic4}.
To overcome catastrophic forgetting, Tariq et al.~\cite{Shahroz3} use few data samples from the source domain during transfer learning. However, in practice, for most pre-trained models, either the source domain data is not available or retaining source domain data may raise privacy concerns. Therefore, to encourage maximum applicability in real-world scenarios, we only use the target domain's data and apply knowledge distillation to learn from the pre-trained model (Teacher).
% Catastrophic forgetting may occur due to training with data from different domains.

\textbf{Pre-processing. }
First of all, we extract the frames ($x$) from real and deepfake videos using a custom code written on top of the FFmpeg library. Then, we use the MTCNN library~\cite{MTCNN} for face landmark detection. The faces are cropped and aligned to the center. We set $128\times128\times3$ to be the resolution of $x$. \textit{Note: Instead of stretching $x$ to match the square aspect ratio (1:1), we crop a bounding box of $128\times128$ from $x$ with the face at the center of the frame. This way, we can avoid stretching the face on the horizontal axis and keep the face in more natural ratio (see Fig.~\ref{fig:motivation})}.

\textbf{Teacher. }
The first step of FReTAL is to train a base model. We will refer to this base model as the pre-trained or teacher model ($f_T$). For the deepfake detection task, this teacher model is trained on the source domain as a binary classifier to distinguish between real and deepfake images (e.g., Pristine and Face2Face). We set the teacher as untrainable for the whole domain adaptation process, and only the student model is trained from the step onwards.

% The first step of FReTAL is to train the pre-trained models, that is the binary classifiers for detecting whether an image is real or fake.
% R\"ossler et al.~\cite{FaceForensics++} demonstrated that 
% In the FaceForensics++~\cite{rossler2018faceforensics} paper, they demonstrated that XceptionNet~\cite{chollet2017xception} showed the state-of-the-art for image classification task while transfer learning.

\textbf{Student. } Once the teacher model is fully trained on the source domain, we copy the weights of teacher model ($f_T$) to the student model ($f_S$). The student model is then trained on the target domain using KD loss and feature-based representation learning. As illustrated in Fig.~\ref{fig:pipeline}, this process is different from the usual teacher-student model.
We provide the details of KD loss and feature-based representation learning in the following sections.

\textbf{Feature-based Representation Learning.}
We assume that similar features must exist between different types of deepfakes. Therefore, a model trained on the source domain (Teacher) can help the student learn the target domain with fewer data samples. Before training on the target domain, the student model is just a copy of the teacher model. Then, we provide both the teacher (untrainable) and student model with the target domain's data to obtain its feature representation ($\Phi_n^t$ for teacher and $\Phi_n^s$ for student), as shown in Figure 2. Instead of storing features of all of the target domain's data, we only store distinguishable features. To do so, we apply softmax to both models' output. 
The softmax function takes as input a vector $v$ of $K$ real numbers, and normalizes it into a probability distribution consisting of $K$ probabilities proportional to the exponentials of the input numbers (i.e., between 0 and 1).
% The softmax function takes a $\mathbb{R}^{M\times N}$real number as input and normalizes them into a probability distribution between 0 and 1. 
Using this output, we create a feature storage from $\lambda_a$ to $\lambda_b$ in $i$ unit intervals, 
% given as $\big\{\lambda_a+k\mid k\in\{0.0,0.1,0.2,\dots,\lambda_b\}\big\}$, %in the range of $\lambda_a$ to $\lambda_b$ with $k$ unit length
as shown in Figure 2. It helps to minimize the domain shifting in the learning process by segmenting the features. As the distribution between real and fake data is different, we store the features of real and fake data separately. We calculate the difference between $\Phi^s$ and $\Phi^t$ using our feature-based square loss $\mathcal{L}_{FSL}$, as follows:
\begin{equation}
\large
    \mathcal{L}_{FSL}=\sum_{\substack{n=\lambda_a\\\text{step }i}}^{\lambda_b} \left \| \Phi_n^s-\Phi_n^t \right \|_2^2
    \label{eq:squareloss}
\end{equation}
\textbf{Domain Adaptation with Knowledge Distillation. }
To reduce the impact of catastrophic forgetting and domain shift, we apply cross-entropy loss and KD loss proposed by Hinton et al.~\cite{hinton2015distilling} while training the student model on the target domain.
% To reduce the ``Catastrophic Forgetting'' from shifting the domain distribution, We apply the cross-entropy loss and KD loss by proposed Hinton et al.~\cite{hinton2015distilling}.
% empirically set \mathcal{\alpha} and Temperature \mathcal{T} are 
Class probabilities are usually generated by neural networks using a softmax output layer that transforms the logit, $x_i$, computed for each class into a probability, $\sigma(x)_i$, by comparing $x_i$ with the other logits $x_j$, as follows:
\begin{equation}
\large
    \sigma(x)_i = \frac{exp(\frac{x_i}{\mathcal{T}})}{\sum_{j=1}^{N} exp(\frac{x_j}{\mathcal{T}}))},
    \label{eg:softmax}
\end{equation}
where $\mathcal{T}$ is the temperature that helps the student model mimic the teacher model by softening the probability distribution over the classes. The softmax function's probability distribution becomes softer by increasing $\mathcal{T}$, revealing which classes the teacher considered to be more similar to the predicted class. In general, KD loss is commonly expressed as minimizing the objective function:
\begin{equation}
\large
    \sum_{x_i\in \mathcal{X}} \mathcal{L}(f_T(x_i),f_S(x_i)),
    \label{eg:KD_general}
\end{equation}
where $x_i$ is the input, $f_T$ is the teacher, $f_S$ is the student, and $\mathcal{L}$ is a loss function that penalizes the difference between teacher and the student. In this work, we use cross-entropy for the loss function $\mathcal{L}$. Therefore, from  Eq.~\eqref{eg:softmax} and~\eqref{eg:KD_general}, we can express our KD loss $\mathcal{L}_{KD}$, as follows:
\begin{equation}
% \large
    \mathcal{L}_{KD}=\sum_{x_i\in \mathcal{X}} \sigma(f_T(x_i,y_i)) \log \sigma(f_S(x_i,\hat y_i))),
    \label{eq:KD_Ours}
\end{equation}
where $\sigma$ is the softmax with temperature, $y_i$ is the output label, and $\hat y_i$ is the output of the teacher $f_T$. In addition to KD loss, we also use cross-entropy loss in our student model $f_S$ given as:
\begin{equation}
% \large
    \mathcal{L}_{CE} = {\sum_{n=1}^{N} y_i\log\sigma(f_{S}(x_i,y_i);\mathcal{T}=1)}
    \label{eq:crossentropy}
\end{equation}
Therefore, the loss function of our Feature  Representation Transfer Adaptation Learning method can be written using Eq.~\eqref{eq:squareloss}, \eqref{eq:KD_Ours}, and \eqref{eq:crossentropy}, as follows:
\begin{equation}
    \mathcal{L}_{FReTAL}= \rho_{1}\mathcal{L}_{FSL}+\rho_{2}\mathcal{L}_{KD}+\rho_{3}\mathcal{L}_{CE},
    \label{eq:FReTALloss}
\end{equation}
where $\rho_{1}$, $\rho_{2}$, and $\rho_{3}$ are scaling factors to control the three loss terms.

\section{Experiment}
\label{sec:Experiment}

We compared FReTAL with several transfer learning methods. In this section, we will describe the implementation details of FReTAL, as well as training and testing details of all detection models.

\subsection{Dataset Description}

To compare our method with several baselines, we used DeepFake (DF), Face2Face (FS), FaceSwap (FS), and NeuralTextures(NT) datasets from FaceForensics++~\cite{rossler2018faceforensics}. The pristine videos from~\cite{rossler2018faceforensics} are used as real videos.
In Table~\ref{tab:dataset_details}, we describe all the datasets used for base training (Teacher) and transfer learning (Student). We used the 750 videos for training the teacher model and only ten videos for training the student model during domain adaptation (or transfer learning for brevity). The remaining 125 videos are used for validation and 125 for testing. In contrast to Tariq et al.~\cite{Shahroz3}, we do not use the source domain dataset during transfer learning.

% To compare our method with several baselines, we used DeepFake (DF), Face2Face (FS), FaceSwap (FS), and NeuralTextures(NT) datasets from FaceForensics++~\cite{rossler2018faceforensics}. The pristine videos from ~\cite{rossler2018faceforensics} are used as real videos.
% In Table~\ref{tab:dataset_details}, we describe all the datasets used for base training (Teacher) and transfer learning (Student). We used the 750 videos for training while base-training, and only 10 videos for training while transfer-learning, and remained 125 for validation and 125 for testing. We note that we proceed without source data through we proposed feature-based presentation learning and output-based KD loss.

\begin{table}[t!]
\centering
\caption{The details of datasets used for training and testing.}
% \vspace{-10pt}
\label{tab:dataset_details}
\resizebox{1\linewidth}{!}{%
\begin{tabular}{lcccc} 
\toprule
\multicolumn{1}{c}{ \textbf{Datasets} } & \begin{tabular}[c]{@{}c@{}}\textbf{Total}\\\textbf{Videos} \end{tabular} & \begin{tabular}[c]{@{}c@{}}\textbf{Training}\\\textbf{Videos} \end{tabular} & \begin{tabular}[c]{@{}c@{}}\textbf{Transfer}\\\textbf{Learning} \end{tabular} & \begin{tabular}[c]{@{}c@{}}\textbf{Testing}\\\textbf{Videos} \end{tabular} \\ 
\hline
Pristine (Real) & 1,000 & 750 & 10 & 250 \\ 
\hline
DeepFake (DF) & 1,000 & 750 & 10 & 250 \\ 
\hline
FaceSwap (FS) & 1,000 & 750 & 10 & 250 \\ 
\hline
Face2Face (F2F) & 1,000 & 750 & 10 & 250 \\ 
\hline
Neural Textures (NT) & 1,000 & 750 & 10 & 250 \\ 
%\hline
%Deepfake Detection (DFD) & 300 & 250 & 10 & 50 \\ 
%\hline
%DeepFake-in-the-Wild (DFW) & 200 & \multicolumn{2}{c|}{\textit{DFW is only used for testing}} & 200 \\
\bottomrule
\end{tabular}
}%\vspace{-12pt}
\end{table}

\subsection{Baselines}
We explored several baselines for comparison. The following is a brief detail on them.
\begin{enumerate}
    \item \textit{\textbf{G{\"u}era et al.}}~\cite{FT_SQ2}:  deployed a stack of CNN on top of an LSTM network to detect deepfake. The CNN module outputs the feature vector fed to the LSTM module that generates the sequence descriptors and passes them to a fully connected layer with softmax to generate probabilities.
    
    \item \textit{\textbf{Sabir et al.}}~\cite{DFD2}:  used DenseNet with a bidirectional RNN to achieve high accuracy on DeepFake, FaceSwap, and Face2Face datasets. 
     
    \item \textit{\textbf{ShallowNet}}: Tariq et al.~\cite{Shahroz2} demonstrated that ShallowNet~\cite{Shahroz1} detects GAN-generated images with high accuracy. We developed ShallowNet using Python and TensorFlow.
     
    \item\textit{\textbf{Xception}}~\cite{chollet2017xception}:  is considered as the state-of-the-art deep learning model for image classification task. Also, R\"ossler et al.~\cite{FaceForensics++} demonstrated that Xception achieves the best accuracy on FaceForensics++ dataset. We used the PyTorch implementation of Xception.
   
\end{enumerate}

The code for CNN+LSTM and DBiRNN are not publicly available; therefore, we implemented them and tried our best to match the original paper’s experimental settings.

\subsection{Domain Adaptation Methods}
In addition to the baselines experiments, we explored several domain adaptation methods as follows:

\begin{enumerate}
    \item \textit{\textbf{FT}}: We apply general transfer learning (fine-tuning) on aforementioned baseline methods without layer freezing.
    
    \item \textit{\textbf{T-GD}}: Jeon et al.~\cite{transferlearning_tgd} propose T-GD that can achieve high performance and prevent the catastrophic forgetting by combining with L2-SP and self-training. We use T-GD to perform transfer learning with Xception model.
    
    \item \textit{\textbf{KD}}: We only use KD loss $\mathcal{L}_{KD}$ component from our $\mathcal{L}_{FReTAL}$ loss function to perform domain adaptation on Xception.
    
\end{enumerate}

\subsection{Implementation Details of FReTAL}
Due to the consistent performance of Xception in many face classification and deepfake detection tasks~\cite{FaceForensics++,Shahroz2,Shahroz1,SAMGAN,transferlearning_tgd}, we select Xception as the backbone model for our FReTAL method. We use the PyTorch implementation of Xception, pre-trained on the ImageNet dataset. We set the value of hyper-parameter values as follows: $\lambda_a=0.5$, $\lambda_b=1.0$, $i=0.1$, $\mathcal{T}=20$, $\rho_{1}=1.0$, $\rho_{2}=1.0$, and $\rho_{3}=1.0$. Therefore, the range for feature storage is $\{(0.5-0.6),(0.6-0.7),\dots,(0.9-1.0)\}$. For training, we used the stochastic gradient descent (SGD) with a learning rate of 0.05 with a momentum of 0.1, and the number of iterations is set to 100. We applied early stopping with a patience of 5. 
%We used the PyTorch implementation of Xception that is pre-trained on the ImageNet dataset.
% To compare our method with several baselines, we fix the parameter setting of FReTAL for the tasks without when transferring the `DeepFake' to the `FaceSwap' case. Also, We use the stochastic gradient descent (SGD\cite{?}) with a learning rate of 0.0001, and the number of iterations is set 100.

% \textbf{Feature-based Representation Learning}
% We assume that there are similar latent features\cite{CITE} between the target and source domain because the task is hardly similar like the deepfake dataset.
% Before transfer learning, we store the feature scalar compressed by inputting the data into the pre-trained teacher model.
% The true positive and true negative data be stored in feature storage, real and fake respectively.
% These features we call ``\mh{representation feature scalar ?}'', which is the information that the student will refer to while transfer learning.
% We apply the softmax function to output data from the teacher model. The reason is to minimize domain shifting in the learning process by segmenting sophisticated presentation feature storage.
% In other words, in the binary classification task such as deepfake detection, we build five storage split from 0.5 to 1.0 range to 0.1 units.
% After that, we take indexing ``Real'' and ``Fake'' to be distinguished. And then, we store in the {$10//numclass$} th storage.

\begin{table}
\centering
\caption{\textbf{Teacher model performance on source dataset (HQ).} Xception performs the best among the baselines. All the results are in percentages (\%) and best are highlighted in bold.}
\label{tab:model_perform}
\label{tab:In-Domain_Attack}
\resizebox{\linewidth}{!}{%
\begin{tabular}{lccccc} 
\toprule
\textbf{Method} & \textbf{DF }  & \textbf{FS}  & \textbf{F2F}  & \textbf{NT}  & \textbf{Avg.} \\ 
\hline
G{\"u}era et al.~\cite{FT_SQ2} & 78.51 & 77.75 & 71.87 & 90.54 & 77.80 \\
Sabir et al.~\cite{DFD2} & 80.54 & 80.56 & 73.12 & 94.38 & 82.21 \\
ShallowNet & 88.97 & 93.33 & 75.26 & 99.45 & 87.08 \\
Xception & \textbf{99.00} & \textbf{99.29} & \textbf{99.26} & \textbf{99.46} & \textbf{99.25} \\
\bottomrule
\end{tabular}
}
\end{table}

\begin{table}
\centering
\caption{\textbf{Teacher model performance on source dataset (LQ) and zero-shot performance.} We are only presenting the results of Xception model on low quality as it is the best performer on HQ dataset. The source dataset results (diagonal) are highlighted in bold.}
\label{tab:LQ_zeroshot}
\resizebox{\linewidth}{!}{%
\begin{tabular}{lcccc} 
\toprule
\textbf{Method} & \textbf{DF (\%)}  & \textbf{F2F (\%)}  & \textbf{FS (\%)}  & \textbf{NT (\%)}  \\ 
\hline
Xception (DF) & \textbf{99.41}  & 56.05 & 49.93 & 66.32 \\ 
\hline
Xception (F2F) & 68.55 & \textbf{98.64}  & 50.55 & 54.81 \\ 
\hline
Xception (FS) & 49.89 & 54.15 & \textbf{98.36}  & 50.74 \\ 
\hline
Xception (NT) & 50.05 & 57.49 & 50.01 & \textbf{99.88}  \\
\bottomrule
\end{tabular}
}
\end{table}

\textbf{Machine Configuration.}
We run our experiments using P100 and TITAN RTX GPUs, with 24 GB of dedicated memory. We use Intel Xeon Gold 6230 CPUs with 8 cores each and 256 GB of RAM. The under-lying OS is Ubuntu 18.04.2 LTS 64 bit. We use PyTorch v1.7.0 with CUDA 11.0 and Python 3.8.

\textbf{Evaluation Metrics.}
We use F$_1$-score metric to evaluate the model performance using 125 real and 125 deepfake test videos.

\textbf{Preprocessing and Data Augmentation. }
We extract 16 samples such that each sample of origin and manipulated video contain five consecutive frames (16×5 = 80 images per video).
To extract the face landmark information from extracted frame, we use multi-task CNN (MTCNN)~\cite{MTCNN}. We apply the following normalization settings using PyTorch Transform: [0.5,0.5,0.5]. We use CutMix~\cite{yun2019cutmix} for data augmentation.
% instead of use cutmix to mix the data of different classes as well as the same class, we apply how to mix the different classes only.

\subsection{Configuring Training Models}
% We set the learning rate 0.0001 and stochastic gradient descent with a momentum of 0.1 as optimizer for base training.
% Also, we do not apply any augmentation technology.
% While training pass, training lasts for 50 epochs unless preempted by early stopping.

\textbf{Teacher Model Training. }
We use any source domain to train the teacher model $f_T$ using 750 real (pristine) and 750 deepfake (e.g., Face2Face) videos. After this process, we set the $f_T$ as untrainable.

\textbf{Student Model Training. }
We initialize the student model $f_S$ by copying the weight from the teacher model. Then, we train $f_S$ on any target domain (e.g., FaceSwap). We do not use any source domain data (e.g., Face2Face) when transfer learning to target domain. To compare general transfer learning, we also train T-GD and KD using the same settings. We perform single source to single target transfer learning using several configuration, as shown in Table~\ref{tab:HQ_transfer_results} and ~\ref{tab:LQ_transfer_results}.

\section{Results}
\label{sec:Results}

\begin{table*}[t!]
\centering
\caption{\textbf{Student model performance on target dataset (HQ).}  We evaluate all datasets with four baselines using four domain adaptation methods. The top-row indicates the ``Source $\to$Target'' dataset. Xception + FReTAL demonstrated the best and most consistent performance. The best results are highlighted in bold. Note: Due to space limitation, we show only a selected Source $\to$Target configurations.}
\label{tab:HQ_transfer_results}
\resizebox{\linewidth}{!}{%
\begin{tabular}{l|c|cccccc} 
\toprule
 \textbf{Method}  & \textbf{Domain}  & \textbf{DF$\rightarrow$F2F (\%)} & \textbf{DF$\rightarrow$FS (\%)} & \textbf{F2F$\rightarrow$DF (\%)} & \textbf{F2F$\rightarrow$FS (\%)} & \textbf{FS$\rightarrow$DF (\%)} & \textbf{FS$\rightarrow$F2F (\%)}\\
\hline
\multirow{3}{*}{\begin{tabular}[c]{@{}l@{}}G{\"u}era et al.~\cite{FT_SQ2}\\~ + \textit{FT} \end{tabular}} & Source & 70.21 & 72.35 & 71.87 & 72.41 & 70.32 & 73.15 \\
 & Target & 50.73 & 52.75 & 52.75 & 63.34 & 50.73 & 66.08 \\
 & \multicolumn{1}{c|}{Avg.} & {60.47} & {62.55} & {62.31} & {67.88} & {60.53} & {69.62} \\ 
\hline
\multirow{3}{*}{\begin{tabular}[c]{@{}l@{}}Sabir et al.~\cite{DFD2}\\~ + \textit{FT} \end{tabular}} & Source & 73.56 & 75.36 & 76.45 & 73.83 & 75.84 & 76.32 \\
 & Target & 59.81 & 55.62 & 55.25 & 51.39 & 50.45 & 55.17 \\
 & {Avg.} & {66.69} & {65.49} & {65.85} & {62.61} & {63.15} & {65.75} \\ 
\hline
\multirow{3}{*}{\begin{tabular}[c]{@{}l@{}}ShallowNet\\~ + \textit{FT} \end{tabular}} & Source & 75.26 & 75.85 & 74.84 & 77.85 & 73.31 & 75.19 \\
 & Target & 55.86 & 50.92 & 58.84 & 42.38 & 53.83 & 50.29 \\
 & {Avg.} & {65.56} & {63.39} & {66.84} & {60.12} & {63.57} & {62.74} \\ 
\hline
\multirow{3}{*}{\begin{tabular}[c]{@{}l@{}}Xception\\~ + \textit{FT} \end{tabular}} & Source & 93.65 & 70.00 & 95.10 & 90.35 & 93.77 & 94.91 \\
 & Target & 84.59 & 55.18 & 91.32 & 55.26 & 86.56 & 83.11 \\
 & {Avg.} & {89.12} & {62.59} & {93.21} & {72.81} & {90.17} & {89.01} \\ 
\hline
\multirow{3}{*}{\begin{tabular}[c]{@{}l@{}}Xception\\~ + \textit{T-GD} \end{tabular}} & Source & 92.96 & 73.92 & 96.89 & 90.42 & 92.55 & 94.85 \\
 & Target & 77.89 & 55.64 & 84.55 & 55.60 & 79.38 & 78.49 \\
 & {Avg.} & {85.43} & {64.78} & {90.72} & {73.01} & {85.97} & {86.67} \\ 
\hline
\multirow{3}{*}{\begin{tabular}[c]{@{}l@{}}Xception\\~ + \textit{KD} \end{tabular}} & Source & 95.58 & 82.77 & 96.91 & 84.57 & 95.65 & 96.28 \\
 & Target & 84.31 & 59.55 & 92.51 & 76.45 & 87.05 & 85.12 \\
 & {Avg.} & {89.95} & {71.16} & \textbf{94.72} & {80.51} & \textbf{91.35} & {90.70} \\ 
\hline
\multirow{3}{*}{\begin{tabular}[c]{@{}l@{}}Xception\\\textbf{~ + FReTAL} \end{tabular}} & Source & 95.68 & 88.60 & 98.09 & 93.36 & 92.57 & 96.41 \\
 & Target & 84.54 & 76.23 & 89.90 & 80.63 & 86.45 & 88.64 \\
 & {Avg.} & \textbf{90.11} & \textbf{82.42} & {94.00} & \textbf{82.00} & {89.51} & \textbf{92.53} \\
\bottomrule
\end{tabular}
}
\end{table*}

\begin{table*}[t!]
\centering
\caption{\textbf{Student model performance on target dataset (LQ).}  We evaluate all datasets with Xception using four domain adaptation methods. The top-row indicates the ``Source $\to$Target'' dataset. Xception + FReTAL demonstrated the best performance for all cases. The best results are highlighted in bold. Note: Due to space limitation, we show only a selected Source $\to$Target configurations.}
\label{tab:LQ_transfer_results}
\resizebox{\linewidth}{!}{%
\begin{tabular}{l|c|cccccc} 
\toprule
 \textbf{Method}  & \textbf{Domain} & \textbf{FS$\rightarrow$F2F (\%)} & \textbf{F2F$\rightarrow$FS (\%)} & \textbf{FS$\rightarrow$DF (\%)} & \textbf{DF$\rightarrow$F2F (\%)} & \textbf{F2F$\rightarrow$NT (\%)} & \textbf{DF$\rightarrow$NT (\%)} \\
\hline
\multirow{3}{*}{\begin{tabular}[c]{@{}l@{}}Xception\\~ + FT \end{tabular}} & Source & 40.93 & 84.78 & 80.56 & 89.84 & 87.12 & 88.29 \\
 & Target & 60.30 & 52.97 & 64.61 & 58.24 & 76.78 & 82.40 \\
 & Avg. & \multicolumn{1}{c}{50.62} & \multicolumn{1}{c}{75.05} & \multicolumn{1}{c}{72.59} & \multicolumn{1}{c}{74.04} & \multicolumn{1}{c}{81.95} & \multicolumn{1}{c}{85.35} \\ 
\hline
\multirow{3}{*}{\begin{tabular}[c]{@{}l@{}}Xception\\~ + T-GD \end{tabular}} & Source & 36.08 & 84.70 & 85.98 & 88.07 & 83.22 & 81.23 \\
 & Target & 56.95 & 52.95 & 55.9 & 49.55 & 52.69 & 67.11 \\
 & Avg. & \multicolumn{1}{c}{46.52} & \multicolumn{1}{c}{68.83} & \multicolumn{1}{c}{70.94} & \multicolumn{1}{c}{68.81} & \multicolumn{1}{c}{67.96} & \multicolumn{1}{c}{74.17} \\ 
\hline
\multirow{3}{*}{\begin{tabular}[c]{@{}l@{}}Xception\\~ + KD \end{tabular}} & Source & 48.07 & 84.84 & 80.48 & 82.59 & 86.07 & 89.61 \\
 & Target & 61.40 & 65.26 & 64.63 & 64.34 & 74.56 & 81.03 \\
 & Avg. & \multicolumn{1}{c}{54.74} & \multicolumn{1}{c}{75.05} & \multicolumn{1}{c}{72.56} & \multicolumn{1}{c}{73.47} & \multicolumn{1}{c}{80.32} & \multicolumn{1}{c}{85.32} \\ 
\hline
\multirow{3}{*}{\begin{tabular}[c]{@{}l@{}}Xception\\\textbf{~ + FReTAL} \end{tabular}} & Source & 81.78 & 82.03 & 85.93 & 91.20 & 82.85 & 90.56 \\
 & Target & 64.45 & 68.79 & 65.78 & 62.09 & 83.87 & 83.38 \\
 & Avg. & \multicolumn{1}{c}{\textbf{73.12}} & \multicolumn{1}{c}{\textbf{75.41}} & \multicolumn{1}{c}{\textbf{75.86}} & \multicolumn{1}{c}{\textbf{76.65}} & \multicolumn{1}{c}{\textbf{83.36}} & \multicolumn{1}{c}{\textbf{86.97}} \\
\bottomrule
\end{tabular}
}
\end{table*}

In this section, we present the results for base training (Teacher) and Transfer learning (Student) on both high- and low-quality datasets.

\subsection{Performance of Teacher}
We evaluate the teacher model $f_T$ using four baseline methods on the high-quality deepfake dataset. As shown in Table~\ref{tab:model_perform}, we find that Xception is the best performer among all baselines across all datasets. This result is consistent with \cite{CLRNet,FaceForensics++}. Therefore, based on this result, we selected Xception as the best candidate to perform further experiments. This time we train Xception on low-quality deepfake datasets and additionally check zero-shot performance. As shown in Table~\ref{tab:LQ_zeroshot}, the model does not perform well except for the source domain. These results are also consistent with the high-quality zero-shot performance result of \cite{Shahroz3}. This result shows that deepfake detectors such as Xception only perform well against the type of deepfakes on which they are trained. Therefore, there is a need for a domain adaptation-based method that can perform well against all kinds of deepfakes.

\subsection{Performance of Student using FReTAL}
Following the same settings is the previous experiment. In this experiment, first, the teacher model is trained on the source domain (HQ), and then we fine-tune (FT) the student using transfer learning to learn the target domain (HQ). Furthermore, we apply T-GD, KD, and our FReTAL on Xception. As shown in Table~\ref{tab:HQ_transfer_results}, Xception with our FReTAL method performs the best in most scenarios except for F2F$\to$DF and FS$\to$DF, where Xception + \textit{KD} demonstrates better performance. As Xception + (domain adaptation method) provides the best performance on high-quality deepfakes, we use it for further experiments with low-quality deepfakes. Now instead of high quality, we use low-quality images for both teacher and student models. As shown in Table~\ref{tab:LQ_transfer_results}, Xception + FReTAL performs the best on all source to target configurations across all datasets. This result demonstrates that our FReTAL is a better domain adaptation method for deepfake detection than the other baselines. The low performance of fine-tuning in some scenarios, such as FS$\to$F2F in Table 7, is due to catastrophic forgetting. In contrast, our FReTAL method shows robustness against catastrophic forgetting.

% \hl{We only compare Xception model on low quality as it is the best performer on HQ dataset.}
% In ~\ref{table}, we show that improving the detection performance of target data while maintaining knowledge of pre-routed models of LQ data as well as HQ data.
% In other words, domain adaptability improves while reducing the catastrophic porting phenomenon. The following sections will discuss the results from different experiments in detail.

\subsection{Ablation Study: Feature Representation}
We perform an ablation study by removing the feature-based representation learning component $\mathcal{L}_{FSL}$ and the student's cross-entropy loss $\mathcal{L}_{CE}$ from our FReTAL method. Without these components, our method becomes similar to the KD. And as shown in Table~\ref{tab:HQ_transfer_results} and~\ref{tab:LQ_transfer_results}, Xception + KD performs worse than Xception + FReTAL in most scenarios, which shows that these components are necessary to achieve better performance.

\section{Discussion}
\label{sec:Discussion}

\textbf{Evaluation of DFDC and CelebDF.}
Recently, more sophisticated deepfake datasets such as DFDC~\cite{DFDC} and CelebDF~\cite{CelebDF} have been released. We plan to include these datasets in our experiment in the future. However, it is important to note that if a deepfake detector fails to perform well on low-quality images of FaceForensics++~\cite{FaceForensics++} dataset, they might also fail on more complex datasets such as DFDC and CelebDF. 

\textbf{Performance on Low-quality Deepfakes.}
The performance on low-quality deepfakes, especially for the transfer learning task, is relatively lower ($<90$\%) than high-quality deepfakes.
It means that there is still a lot of room for improvement in domain adaptation for low-quality deepfake detection. We believe that applying the super-resolution method as a data augmentation method on low-quality deepfakes may reduce this gap.

\textbf{Mixed LQ and HQ Deepfake Detection.}
As we know that, models trained on high-quality deepfakes do not perform well on low-quality deepfakes. However, it is interesting to note that the model trained on low-quality deepfakes does not perform well on high-quality deepfakes as well unless we apply the same compression on the high-quality deepfakes to convert them into low-quality. Therefore, programmatically identifying the quality of deepfake is another venue of research. We also want to focus on detecting mixed quality deepfake datasets like DFDC.

\textbf{Limitations and Future Work.}
Detecting talking head types of deepfakes~\cite{samsungtalkinghead} are not explored in this work. Also, recently, full-body gesture-based deepfakes have emerged~\cite{FirstOrderMotion}. It would be interesting to see how FReTAL can be generalized against talking head and full-body deepfakes.
For data collection, it is becoming challenging to distinguish deepfake videos visually. Furthermore, it is not easy to use or download these deepfake videos to train a deepfake detector due to privacy and copyright issues. Therefore, using a minimum amount of freely available data to achieve high performance is preferable in such scenarios. To solve these problems, we will explore an augmentation method for few-shot learning to improve practicality and performance with very few videos or images. Furthermore, we will utilize our feature-based representation learning framework to improving the domain adaptability and generalization capabilities of other deepfake detectors. Future work also includes exploring alternative training strategies that can help improve performance and multi-domain adaptation.

\section{Conclusion}
\label{sec:Conclusion}

% \hl{write conclusion}
Performing domain adaptation for detecting deepfakes is becoming more challenging for low-quality images than high-quality ones. We find that similar features exist between the source and target dataset that can help in domain adaptation. Therefore, we propose a domain adaptation method using feature storage and KD loss in a teacher-student network, where the teacher is not trained on the target domain. Moreover, we demonstrate that applying KD loss without even using the source dataset can reduce catastrophic forgetting, i.e., domain shifting in deepfake detection tasks. We show that by using FReTAL, we can quickly adapt to new types of deepfakes with a reasonable performance using as low as ten samples of the target domain.
For future work, we plan to explore more augmentation methods on the target domain data to improve practicality and performance. We will also utilize our FReTAL framework to improve domain adaptability and generalization capabilities of other deepfake detection models such as CLRNet and Mesonet. 
\section*{Acknowledgment}
This work was partly supported by Institute of Information \& communications Technology Planning \& Evaluation (IITP) grant funded by the Korea government (MSIT) (No.2019-0-00421, AI Graduate School Support Program (Sungkyunkwan University)), (No. 2019-0-01343, Regional strategic industry convergence security core talent training business) and the Basic Science Research Program through National Research Foundation of Korea (NRF) grant funded by Korea government MSIT (No. 2020R1C1C1006004). Additionally, this research was partly supported by IITP grant funded by the Korea government MSIT (No. 2021-0-00017, Original Technology Development of Artificial Intelligence Industry) and was partly supported by the Korea government MSIT, under the High-Potential Individuals Global Training Program (2019-0-01579) supervised by the IITP.

{\small
\bibliographystyle{ieee_fullname}
\bibliography{references}
}

\end{document}